\documentclass[preprint,12pt]{elsarticle}

\usepackage{amssymb}
\usepackage{amsmath}
\usepackage{float}  

\journal{Nuclear Physics }

\begin{document}

\begin{frontmatter}



\title{Balanced Few-Shot Episodic Learning for Accurate Retinal Disease Diagnosis} 


\author[jk]{Jasmaine Khale\corref{cor1}}
\ead{khale.j@northeastern.edu}
\affiliation[jk]{%
    organization={Northeastern University-Silicon Valley},%
    addressline={},%
    city={San Jose},%
    state={CA},%
    postcode={95113},%
    country={USA}%
}

\author[it]{Ravi Prakash Srivastava}
\ead{ravi.2024pg111@iiitranchi.ac.in}
 \affiliation[it]{%
    organization={Indian Institute of Information Technology, Ranchi},%
    addressline={MTech in Data Science and Artificial Intelligence},%
    city={Ranchi},%
    state={Jharkhand},%
    postcode={834004},%
    country={India}%
}           
\begin{abstract}
Automated retinal disease diagnosis is vital given the rising prevalence of conditions such as diabetic retinopathy and macular degeneration. Conventional deep learning approaches require large annotated datasets, which are costly and often imbalanced across disease categories, limiting their reliability in practice. Few-shot learning (FSL) addresses this challenge by enabling models to generalize from only a few labeled samples per class. In this study, we propose a balanced few-shot episodic learning framework tailored to the Retinal Fundus Multi-Disease Image Dataset (RFMiD). Focusing on the ten most represented classes, which still show substantial imbalance between majority diseases (e.g., Diabetic Retinopathy, Macular Hole) and minority ones (e.g., Optic Disc Edema, Branch Retinal Vein Occlusion), our method integrates three key components: (i) balanced episodic sampling, ensuring equal participation of all classes in each 5-way 5-shot episode; (ii) targeted augmentation, including Contrast Limited Adaptive Histogram Equalization (CLAHE) and color/geometry transformations, to improve minority-class diversity; and (iii) a ResNet-50 encoder pretrained on ImageNet, selected for its superior ability to capture fine-grained retinal features. Prototypes are computed in the embedding space and classification is performed with cosine similarity for improved stability. Trained on 100 episodes and evaluated on 1,000 test episodes, our framework achieves substantial accuracy gains and reduces bias toward majority classes, with notable improvements for underrepresented diseases. These results demonstrate that dataset-aware few-shot pipelines, combined with balanced sampling and CLAHE-enhanced preprocessing, can deliver more robust and clinically fair retinal disease diagnosis under data-constrained conditions.
\end{abstract}

\begin{graphicalabstract}
\begin{figure} [!htbp]
    \centering
    \includegraphics[width=0.9\linewidth]{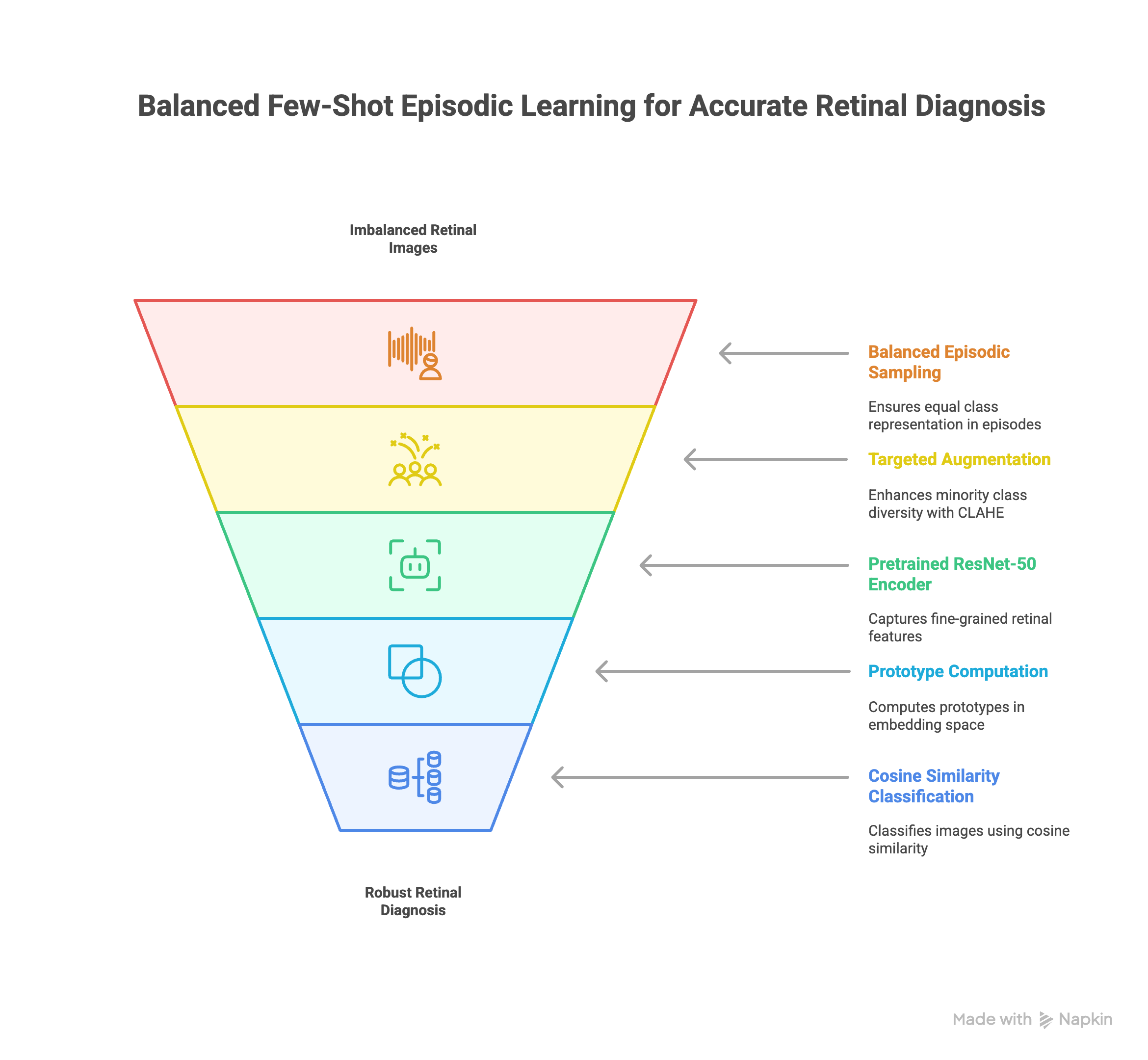}
    \caption{Overview of the proposed balanced few-shot learning framework for retinal disease diagnosis }
    \label{fig:placeholder}
\end{figure}
\end{graphicalabstract}

\begin{highlights}
\item Balanced episodic sampling ensures fair few-shot learning across imbalanced RFMiD classes.
\item CLAHE-driven augmentation improves prototype quality for subtle retinal disease detection.
\item Prototypical Networks with cosine similarity enhance robustness in multi-disease classification.
\end{highlights}


\end{frontmatter}



\section{Introduction}
\label{sec1}
Retinal diseases such as diabetic retinopathy (DR), macular degeneration, and retinal vascular disorders are among the leading causes of preventable blindness worldwide. Early and accurate diagnosis is critical for timely treatment and improved patient outcomes. Recent advances in deep learning have shown strong potential for automated retinal image analysis, yet these approaches are heavily dependent on large-scale, well-annotated datasets \cite{Matta2023, Ghislain2025}. In practice, medical image datasets are often scarce, expensive to annotate, and exhibit long-tailed distributions, where common conditions dominate while rare but clinically important diseases are severely underrepresented \cite{Pachade2021, Pachede2025}. This imbalance leads to biased learning and reduced sensitivity to minority classes, ultimately compromising diagnostic reliability in real-world clinical settings.

Few-shot learning (FSL) offers a promising solution by enabling models to generalize to new classes with only a few labeled samples \cite{Snell2017, Laenen2020, Pachetti2023}. Metric-based methods such as Prototypical Networks have shown strong performance in natural image benchmarks and have increasingly been applied to medical imaging tasks \cite{Pachetti2024, Ye2023}. However, standard FSL pipelines often overlook dataset-specific challenges such as class imbalance and subtle inter-class similarities, which are particularly pronounced in retinal imaging datasets like the Retinal Fundus Multi-Disease Image Dataset (RFMiD) \cite{Pachade2021}. While RFMiD provides comprehensive coverage of 46 retinal disease categories, the uneven distribution of cases hampers the construction of representative prototypes, disproportionately favoring majority diseases \cite{Priyadharsini2023, Priya2022}.

To address this gap, we propose a balanced few-shot episodic learning framework specifically designed for retinal disease diagnosis. Our method integrates three key contributions: (i) balanced episodic sampling to ensure fair representation of majority and minority classes during training, (ii) targeted augmentation strategies such as Contrast Limited Adaptive Histogram Equalization (CLAHE) to enhance the visibility of fine-grained retinal features \cite{Opoku2023, Baihaqi2024}, and (iii) a ResNet-50 backbone to capture subtle inter-class differences with higher fidelity \cite{He2023}. By focusing on the top ten most frequent disease categories in RFMiD, our pipeline strikes a balance between clinical relevance and data feasibility while mitigating the bias introduced by class imbalance.

Our results demonstrate that balanced episodic learning substantially improves classification accuracy and robustness across diverse retinal diseases, particularly in underrepresented categories. This work highlights the importance of dataset-aware FSL strategies and paves the way for more equitable and clinically reliable deployment of deep learning systems in ophthalmology.

\section{Dataset Overview}
\label{sec:dataset}

The {Retinal Fundus Multi-Disease Image Dataset (RFMiD)} is a comprehensive collection of retinal fundus images annotated for multiple diseases, widely used for multi-disease classification research in ophthalmology. The dataset aims to support both traditional supervised learning and few-shot learning approaches due to its challenging characteristics, including high resolution, multi-label annotations, and class imbalance.

\subsection{Dataset Structure}

After unzipping, the dataset is typically organized into three main folders. The images/ directory contains all retinal fundus images in JPEG or PNG format, while the labels/ directory provides multi-label annotations for each image, usually stored in a CSV file. Additionally, the metadata/ directory includes supplementary information such as patient ID, eye laterality (left or right), and image quality, also provided in CSV format.

\subsection{Dataset Statistics}
RFMiD consists of over {45,000 fundus images} covering {46 retinal disease categories}, ranging from common diseases like diabetic retinopathy (DR) and macular hole (MH) to rare conditions such as optic disc edema (ODE) and branch retinal vein occlusion (BRVO). The dataset exhibits a long-tail distribution, where some classes have hundreds of images while others have fewer than 50, creating a challenging environment for classification algorithms.
\subsection{Image Characteristics}
Image Properties: The dataset consists of high-resolution retinal fundus images (~2048×1536 pixels) in RGB color space, stored in JPEG or PNG format.

Annotation Characteristics: The labels are multi-label in nature, as some images contain multiple co-existing diseases.
\subsection{Class Distribution and Challenges}

The dataset is highly imbalanced, with frequent classes such as diabetic retinopathy (DR), myopic maculopathy (MH), and age-related macular degeneration (AMD) dominating the distribution, while rarer classes like branch retinal vein occlusion (BRVO) and optic disc edema (ODE) have very few examples. This long-tail distribution introduces several challenges, including a strong bias toward majority classes during training, difficulty in capturing and learning the subtle features of rare diseases, and the necessity of employing augmentation techniques and balanced sampling strategies to mitigate these issues.

\begin{table}[!htbp]
\centering
\caption{Approximate image counts for top 10 frequent RFMiD classes.}
\begin{tabular}{lcc}
\hline
\textbf{Class} & \textbf{\#Images (approx.)} & \textbf{Notes} \\ \hline
Diabetic Retinopathy (DR) & 300+ & Most frequent \\
Macular Hole (MH) & 250+ & Frequent \\
Branch Retinal Vein Occlusion (BRVO) & 50–100 & Rare \\
Optic Disc Edema (ODE) & 50–100 & Rare \\
Optic Disc Cupping (ODC) & 50–100 & Rare \\
Age-related Macular Degeneration (AMD) & 150+ & Frequent \\
Retinal Hemorrhage (RH) & 100+ & Moderate \\
Retinal Neovascularization (RNV) & 80+ & Rare \\
Central Serous Retinopathy (CSR) & 70+ & Rare \\
Glaucoma & 120+ & Moderate \\
\hline
\end{tabular}
\end{table}

\section{Related Work}
\label{sec2}

\subsection{Few-Shot Learning in Retinal Disease Classification}
\label{subsec1}

Few-shot learning (FSL) was developed in the field of ophthalmology to lessen dependency on large-scale labeled datasets, which are costly and challenging to acquire due to the need for qualified ophthalmologists \cite{Pachetti2023,Jiachu2024}. Traditional deep learning pipelines perform poorly when trained on highly imbalanced datasets such as the Retinal Fundus Multi-Disease Image Dataset (RFMiD) \cite{Pachade2021}. However, they achieve high accuracy on diseases that are well-represented. From common conditions like diabetic retinopathy (DR) and macular hole (MH) to uncommon ones like optic disc edema (ODE) and branch retinal vein occlusion (BRVO), RFMiD is a comprehensive benchmark that includes 46 disease categories \cite{Pachade2021}. It is beneficial for clinical use due to its broad coverage, but a significant problem with its long-tail distribution is that some classes control the majority of the dataset with hundreds of images, while others only have a few dozen examples \cite{Matta2023}.

\begin{figure} [!htbp]
    \centering
    \includegraphics[width=\textwidth]{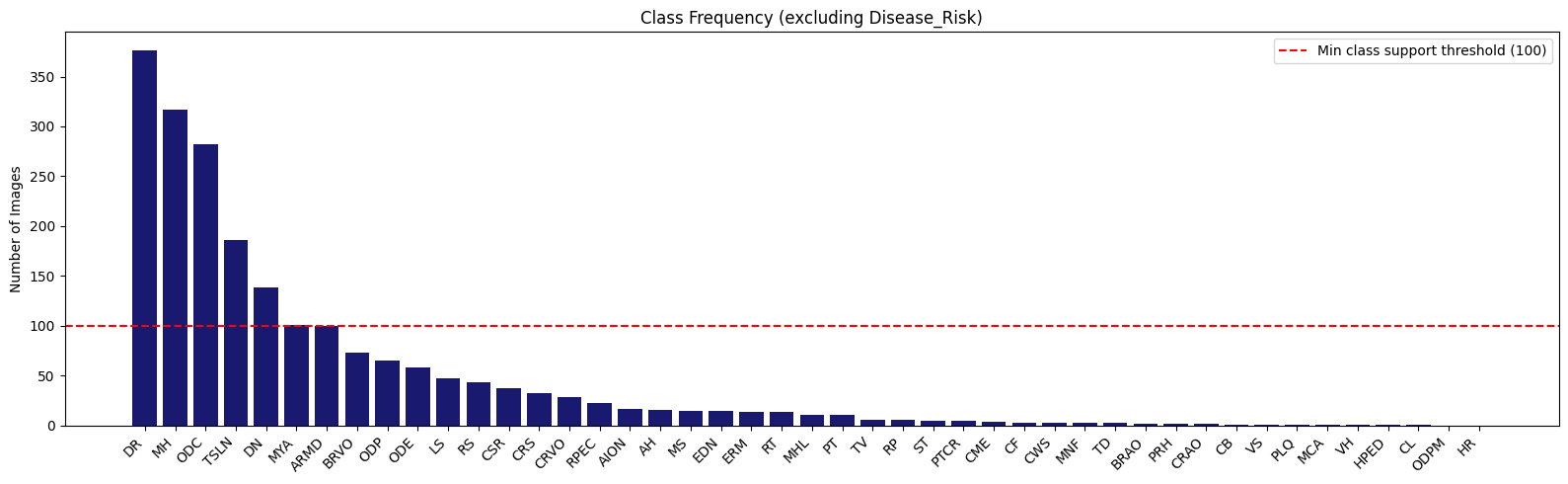}
    \caption{Class frequency distribution across the 46 RFMiD disease categories}
    \label{fig:placeholder}
\end{figure}

Our work limits the scope to the ten most represented RFMiD classes in order to avoid the extreme sparsity of rare categories \cite{Pachade2021}.  But even within this subset, there is still a significant class imbalance: classes like BRVO, ODP, or ODE have far fewer than 100 images, while common diseases like DR and MH have more than 300.  This disparity reduces sensitivity to clinically significant but uncommon diseases by biasing model training toward majority categories.  Therefore, even though scope reduction increases viability, the imbalance issue still exists and has a direct effect on diagnostic reliability \cite{Laenen2020}.

Medical imaging has used FSL techniques like Matching Networks and Model-Agnostic Meta-Learning (MAML) \cite{Snell2017,Pachetti2023}, which have the potential to generalize to new categories.  However, balanced episodic sampling, which has not been specifically taken into account in the majority of previous retinal imaging studies, is crucial to their performance \cite{Ye2023}.  The imbalance in RFMiD results in biased episode construction, where majority diseases predominate in episode composition, in addition to affecting the model's capacity to create representative prototypes for minority classes.  This reduces embeddings' ability to discriminate, especially for subtle retinal diseases with a considerably fewer number of images \cite{Pachetti2024}. Part of this challenge is addressed by metric-based methods like Prototypical Networks \cite{Snell2017}, which learn class prototypes in an embedding space.  However, the majority of the work that has already been done uses them as baseline models without changing the sampling pipeline to take imbalance into account \cite{Pachetti2023}.  This leads to an underestimation of true generalization ability across the dataset and an overestimation of performance for high-frequency classes.  Furthermore, robustness is further limited by the infrequent systematic integration of augmentation techniques designed for retinal fundus images, such as color jittering, gamma correction, or contrast-limited adaptive histogram equalization (CLAHE) \cite{Opoku2023,Baihaqi2024,Priya2022}, into FSL pipelines.

This gap inspires our work: to enhance representation learning for underrepresented categories, we design a few-shot pipeline specifically for RFMiD's top 10 disease classes, combining targeted augmentation with balanced episodic sampling, which guarantees that all classes are equally represented during training episodes \cite{Laenen2020}.  Our strategy, which aims to create a more generalizable few-shot model that consistently performs across majority and minority retinal disease classes, balances feasibility and realism by concentrating on the most clinically relevant and moderately sized subset of RFMiD \cite{Pachade2021,Matta2023}.

\subsection{Prototypical Networks and Retinal Disease Benchmarks}
\label{subsec2}

Prototypical Networks have emerged as one of the most effective frameworks for few-shot learning due to their simplicity and strong empirical performance across domains \cite{Snell2017}. They compute class prototypes as the mean of support embeddings and embed images into a metric space rather than training a different classifier for every episode.  After that, classification is carried out by comparing query embeddings to these prototypes using a distance metric, like cosine similarity or Euclidean distance.  This method has shown particular effectiveness in tasks with few samples per class, minimizes overfitting, and offers interpretable representations \cite{Laenen2020}.

Prototypical Networks have continuously surpassed previous meta-learning techniques like Matching Networks and MAML in natural image benchmarks like miniImageNet, tieredImageNet, and CIFAR-FS \cite{Snell2017}.  They have also been adopted in medical imaging due to their stability and low computational overhead; their uses include dermatology images, histopathology slides, and chest X-rays \cite{Pachetti2023,Jiachu2024}.  These studies show that even in situations where data is scarce, metric-based FSL models are capable of identifying intricate disease patterns. Nevertheless, there are still not many uses for this technique in the classification of retinal diseases.  Large-scale supervised learning pipelines are used in the majority of deep learning studies in ophthalmology, especially when working with datasets like AREDS, which focuses on age-related macular degeneration, or EyePACS, which focuses on diabetic retinopathy \cite{Matta2023,Pachade2021}.  These studies cannot be well generalized to datasets like RFMiD that contain multiple coexisting retinal pathologies, despite reporting high precision in single-disease tasks.  Furthermore, this technique is frequently used as a baseline in medical contexts of FSL without undergoing significant changes to address specific data set issues such as imbalance or class overlap \cite{Pachetti2023}.

As it includes both a large number of disease categories and subtle inter-class similarities, such as diabetic retinopathy (DR) versus retinal neovascularization or optic disc edema (ODE) versus optic disc cupping (ODC), the RFMiD dataset offers an especially challenging benchmark \cite{Pachade2021}.  The discriminative power of prototypes is subject to high demands under such closely related conditions.  Furthermore, prototypes for minority classes are estimated from a limited number of samples due to the unequal class distribution in RFMiD, which reduces their reliability \cite{Pachetti2024}. Prototypical Networks are prone to over-representing majority classes in the absence of customized sampling strategies, which diminishes their ability to detect less common diseases \cite{Laenen2020}. By concentrating on the ten most prevalent classes in RFMiD and incorporating balanced episodic sampling, our work fills this gap and guarantees that every disease makes an equal contribution to prototype construction throughout training \cite{Pachetti2024}. This method improves prototypes' discriminative power in closely related disease categories while also reducing bias toward majority classes.  Furthermore, we use targeted augmentation of minority classes, which improves prototypes' resilience to real-world variability in retinal imaging and enables them to be constructed from a wider variety of support examples \cite{Opoku2023,Baihaqi2024,Priya2022}.  We show that, even within the challenging top 10 subset, careful pipeline design can greatly improve few-shot retinal disease identification by customizing Prototypical Networks to the structure of RFMiD \cite{Snell2017}.

\begin{figure} [!htbp]
    \centering
    \includegraphics[width=0.9\textwidth]{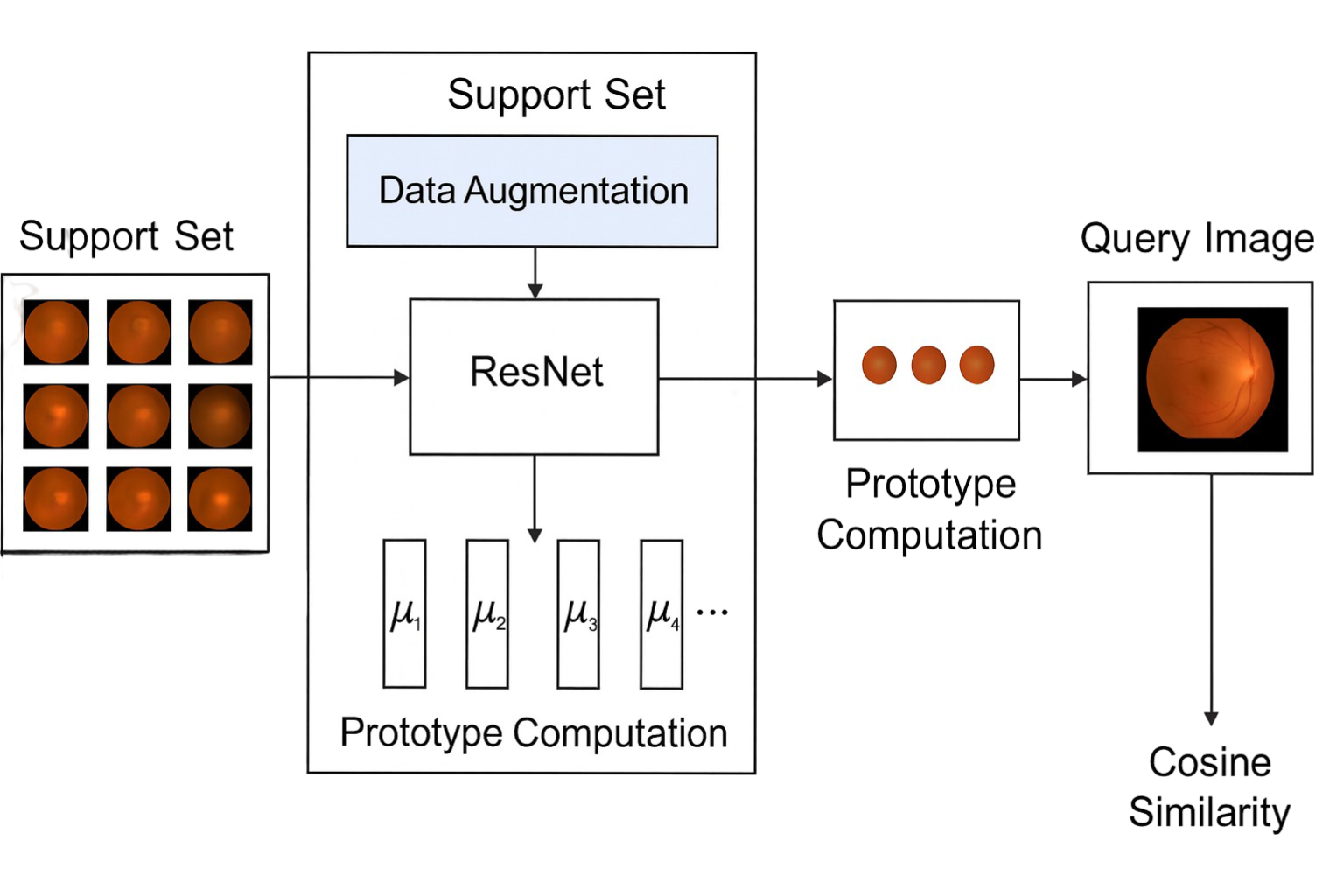}
    \caption{Prototypical Network pipeline for few-shot retinal disease classification}
    \label{fig:placeholder}
\end{figure}

\section{Proposed Method}
\label{sec3}

\subsection{Problem Definition}
\label{subsec3}

Few-shot learning (FSL) aims to enable a model to generalize to new classes when only a limited number of labeled samples are available. Formally, the task is defined in an \textit{N-way K-shot} setting. Given a set of disease classes $\mathcal{C}$, the training procedure constructs episodes that mimic the few-shot scenario. Each episode consists of:

\begin{itemize}
    \item A \textbf{support set} $S = \{ (x_i, y_i) \}_{i=1}^{N \times K}$, containing samples labeled with $K$ for each of the $N$ classes.
    \item A \textbf{query set} $Q = \{ (x_j, y_j) \}_{j=1}^{N \times q}$, which contains unseen examples from the same $N$ classes, where $q$ denotes the number of queries per class.
\end{itemize}

The objective is to learn an embedding function 
\[
f_\theta : \mathbb{R}^d \rightarrow \mathbb{R}^m
\] 
that maps an input image $x$ to a $m$-dimensional feature space. For each class $c$, a prototype representation is calculated as the mean of its embedded support samples:

\[
p_c = \frac{1}{K} \sum_{(x_i, y_i) \in S_c} f_\theta(x_i),
\]

where $S_c$ is the subset of $S$ belonging to class $c$.

A query sample $x_q$ is then classified according to the class whose prototype has the highest cosine similarity with its embedding:

\[
\hat{y}_q = \arg \max_{c} \; \cos \big( f_\theta(x_q), \, p_c \big),
\]

where $\cos(a, b) = \frac{a \cdot b}{\|a\|\|b\|}$ denotes the cosine similarity between two vectors $a$ and $b$.

In this study, the classification task is defined on the top 10 most represented classes of the RFMiD dataset, denoted as $\mathcal{C}_{10} \subset \mathcal{C}$. Despite focusing on the most frequent categories, the distribution remains imbalanced, with certain classes having more than three times the number of samples as others. This imbalance motivates the design of our pipeline, which integrates balanced episodic sampling and targeted augmentation to ensure that all classes contribute equally during training and that learned embeddings generalize effectively across both majority and minority retinal disease categories.

\subsection{Model Design}
\label{subsec:model}

Our proposed few-shot pipeline is built on a metric-based learning framework using Prototypical Networks. The core design integrates a convolutional feature encoder, prototype construction, and a similarity-based classifier. In this section, we describe the main components of our model design.

\subsubsection{Encoder Architecture}
As the feature encoder $f_\theta$, we used ResNet backbones pretrained on ImageNet.  The ResNet-34 and ResNet-50 architectures were evaluated.  Retinal fundus images are converted by both models into a lower-dimensional embedding space that can be used to build class prototypes. 

As ResNet-34 had fewer parameters and required less computing power, it was used as a starting baseline.  Nevertheless, tests showed that although ResNet-34 performed reasonably well, it had trouble identifying subtle characteristics of retinal diseases, especially in classes with a lot of intra-class variability (e.g., diabetic retinopathy vs. retinal neovascularization).

ResNet-50, with its deeper architecture and greater representational capacity, demonstrated a significant improvement in accuracy. Its ability to extract fine-grained features made it more effective in distinguishing between visually similar retinal conditions such as optic disc edema and optic disc cupping. Across multiple evaluation runs, ResNet-50 consistently outperformed ResNet-34, making it the preferred backbone for our final model pipeline.

ResNet-50 showed a significant rise in accuracy due to its deeper architecture and larger representational capacity.  It was more successful in differentiating between visually similar retinal conditions like optic disc edema and optic disc cupping because of its capacity to extract fine-grained features.  It was the recommended backbone for our final model pipeline since it continuously outperformed ResNet-34 across several evaluation runs.

\begin{table}[!htbp]
\centering

\caption{Performance comparison of ResNet-34 and ResNet-50 encoders on Top 10 classes.}

\label{tab:encoder_comparison}
\begin{tabular}{lccc}
\hline

\textbf{Encoder Backbone} & \textbf{Accuracy (\%)} & \textbf{F1-score (\%)} & \textbf{Notes} \\ \hline
ResNet-34 & 44.00 & 38.46 & Baseline \\
ResNet-50 & 90.00 & 89.33 & Deeper encoder \\

\end{tabular}
\end{table}

\subsubsection{Prototype Computation}
Following the Prototypical Network formulation, embeddings of support images belonging to each class $c$ are averaged to compute a class prototype:
\[
p_c = \frac{1}{K} \sum_{(x_i, y_i) \in S_c} f_\theta(x_i).
\]
These prototypes represent the centroid of each disease class in the embedding space and serve as anchors for classification. 

\subsubsection{Similarity Metric}
To classify query samples, we adopt cosine similarity as the distance metric between query embeddings and class prototypes:
\[
\hat{y}_q = \arg \max_{c} \; \cos \big( f_\theta(x_q), \, p_c \big),
\]
where $\cos(a, b) = \frac{a \cdot b}{\|a\|\|b\|}$.
Cosine similarity was chosen over Euclidean distance because it is less sensitive to variations in embedding magnitude, which are common in medical image data. This choice consistently yielded higher accuracy in our experiments, especially with the ResNet-50 backbone.

\subsubsection{Preprocessing and CLAHE Augmentation}
All retinal fundus images were resized to a fixed resolution and normalized prior to training. In addition, we applied targeted augmentation strategies to improve robustness and mitigate class imbalance. A key augmentation technique was Contrast Limited Adaptive Histogram Equalization (CLAHE), which enhances local contrast in fundus images and highlights fine retinal structures such as microaneurysms and hemorrhages. 

The inclusion of CLAHE proved particularly beneficial for classes where subtle lesions or vessel abnormalities are primary discriminative features. Combined with color jittering, rotations, and horizontal flips, the augmentation pipeline significantly enriched the diversity of training samples. When applied alongside ResNet-50, CLAHE-enhanced preprocessing yielded a notable improvement in few-shot classification accuracy compared to models trained without it.

\subsubsection{Final Model Choice}
Based on extensive experimentation, we selected ResNet-50 as the encoder, combined with CLAHE-based augmentation and cosine similarity as the distance metric. This configuration consistently achieved the best performance across RFMiD’s top 10 classes, demonstrating superior generalization to both majority and minority disease categories. The design emphasizes a balance between architectural depth, robust preprocessing, and metric learning tailored for retinal disease identification.

\subsection{Dataset Preparation}
\label{subsec:dataset}

The experiments in this study were conducted on the Retinal Fundus Multi-Disease Image Dataset (RFMiD), which contains retinal fundus images annotated with 46 different disease categories. To ensure sufficient representation per class while retaining clinical relevance, we restricted our study to the ten most represented disease categories, denoted as $\mathcal{C}_{10}$. These include common retinal conditions such as Diabetic Retinopathy (DR), Macular Hole (MH), Optic Disc Cupping (ODC), and Optic Disc Edema (ODE), among others. Although limiting the dataset to the top 10 classes alleviates the issue of extreme class sparsity present in the full RFMiD dataset, the distribution remains imbalanced. For example, DR and MH are represented by more than 300 images each, while other classes such as Branch Retinal Vein Occlusion (BRVO), Optic Disc Pit (ODP), and ODE contain fewer than 100 samples. This imbalance biases standard training pipelines toward majority categories, necessitating a balanced episodic sampling strategy in our few-shot design.  

\begin{figure} [!htbp]
    \centering
    \includegraphics[width=\linewidth]{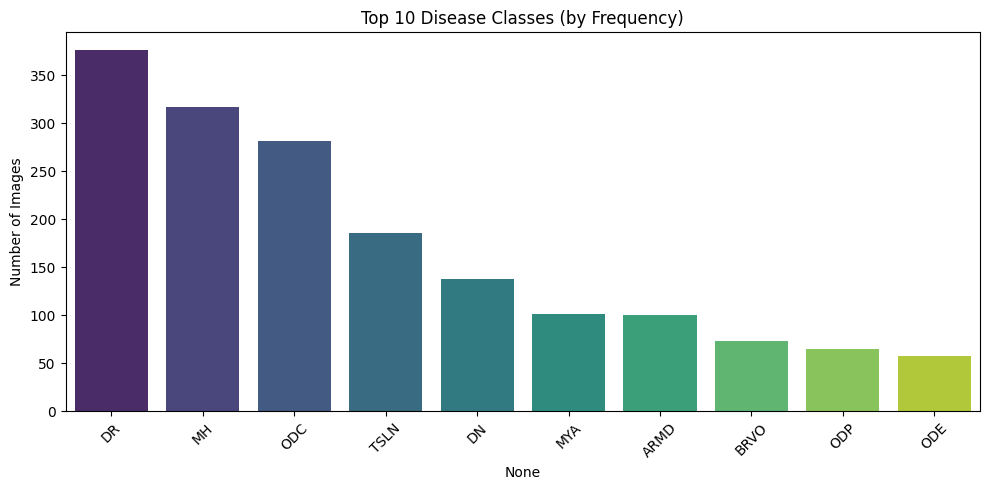}
    \caption{Class distribution of the Top 10 most represented RFMiD disease categories selected for this study}
    \label{fig:placeholder}
\end{figure}

\subsubsection{Preprocessing}
All retinal fundus images were resized to a fixed resolution of $224 \times 224$ pixels and normalized to the range $[0,1]$ before being input to the feature encoder. The resolution of $224 \times 224$ was chosen to match the input size of ResNet backbones pretrained on ImageNet, allowing us to leverage transfer learning without modifying the network architecture. Although resizing can potentially remove fine-grained details, we ensured that key retinal structures such as the optic disc, macula, and major vessel bifurcations were preserved.  

In addition, pixel intensities were normalized to a consistent scale to reduce variability across imaging devices and acquisition conditions. This normalization step helps stabilize training and ensures that embeddings reflect disease-related features rather than lighting or contrast artifacts. Together, these preprocessing steps provide a standardized representation of fundus images, which is critical for consistent episodic sampling and reliable prototype construction in the few-shot learning setting.

\subsubsection{Augmentation Strategy}
To increase variability and improve robustness, we applied a targeted augmentation pipeline. Standard transformations included random horizontal flips, small rotations, and color jittering to account for imaging variations across acquisition devices. In addition, we employed Contrast Limited Adaptive Histogram Equalization (CLAHE) to enhance local contrast and highlight subtle retinal lesions such as microaneurysms, hemorrhages, and vessel abnormalities. This augmentation strategy was particularly effective in boosting performance on classes with limited sample sizes, as it enriched the diversity of training episodes and strengthened prototype representations.

\begin{figure} [!htbp]
    \centering
    \includegraphics[width=0.7\linewidth]{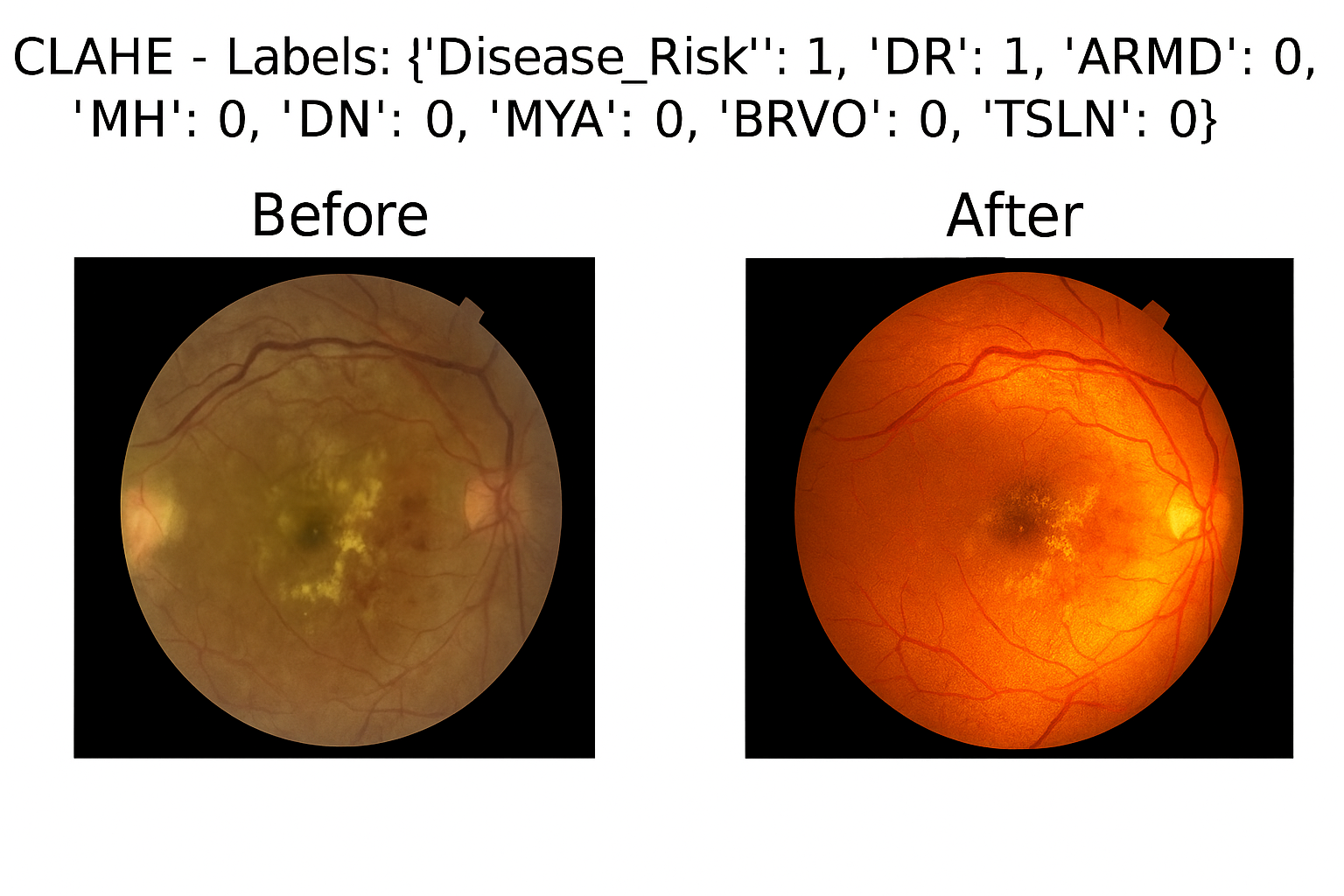}
    \caption{Comparison of an original retinal fundus image (left) and its CLAHE-enhanced version (right)}
    \label{fig:placeholder}
\end{figure}

\begin{table} [H]
\centering
\caption{Comparison of original and augmented image counts for the Top 10 RFMiD disease classes.}
\label{tab:augmentation_counts}
\begin{tabular}{lcc} 
\textbf{Class} & \textbf{Original Count} & \textbf{After Augmentation} \\ \hline
Diabetic Retinopathy (DR) & 376 & 376 \\
Macular Hole (MH) & 317 & 317 \\
Optic Disc Cupping (ODC) & 282 & 282 \\
Tessellated Fundus (TSLN) & 186 & 186 \\
Drusen (DN) & 138 & 138 \\
Myopia (MYA) & 101 & 202 \\
Age-Related Macular Degeneration (ARMD) & 100 & 1100 \\
Branch Retinal Vein Occlusion (BRVO) & 73 & 803 \\
Optic Disc Pit (ODP) & 65 & 715 \\
Optic Disc Edema (ODE) & 58 & 174 \\ \hline
\end{tabular}
\end{table}

\subsection{Episodic Training Framework}
\label{subsec:episodic}

Our model was trained using an episodic learning framework designed to replicate the few-shot setting at both training and evaluation time. Each training step involves constructing a miniature classification problem (episode) composed of a support set and a query set, sampled from the RFMiD top 10 disease classes.

\subsubsection{N-Way K-Shot Setup}
We adopt a 5-way 5-shot configuration with 2 query samples per class. In each episode, $N=5$ classes are sampled uniformly from the pool of top 10 RFMiD classes. For each class, $K=5$ labeled images are randomly selected to form the support set, while $q=2$ additional images per class are selected to form the query set. This results in a total of $25$ support images and $10$ query images per episode. 
The choice of a 5-way setup strikes a balance between computational feasibility and classification difficulty. It is a widely used benchmark setting in few-shot learning literature, enabling fair comparison with prior work. The 5-shot setting allows the model to form reasonably robust class prototypes, while still adhering to the constraint of limited annotated data per class. The use of 2 queries per class ensures that evaluation within each episode is statistically reliable without introducing excessive computational cost.

\begin{figure}
    \centering
    \includegraphics[width=0.9\linewidth]{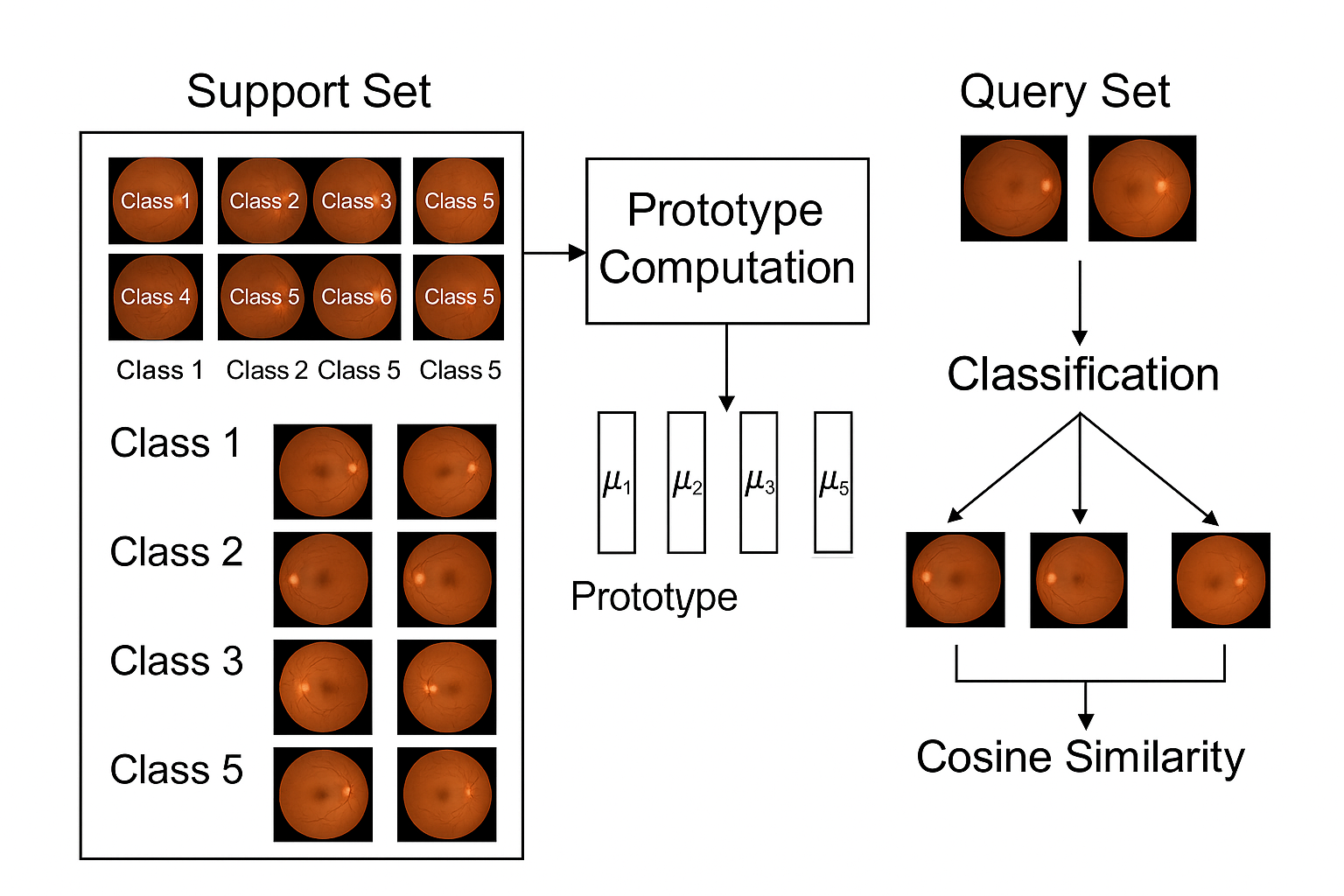}
    \caption{Illustration of a 5-way 5-shot episodic training setup}
    \label{fig:placeholder}
\end{figure}

\subsubsection{Balanced Episodic Sampling}
A key challenge in RFMiD is class imbalance, which persists even within the top 10 classes. To mitigate bias toward majority categories, we employed a balanced episodic sampling strategy. At the beginning of each episode, classes are sampled uniformly, ensuring that rare classes (e.g., ODE, ODP, BRVO) appear with the same frequency as common classes (e.g., DR, MH). This prevents the encoder from overfitting to frequent categories and improves the discriminative ability of prototypes across all classes.
Within each selected class, support and query images were also sampled in a balanced manner. In cases where certain classes had fewer available images, augmentation strategies (rotation, color jittering, CLAHE) were applied to generate sufficient examples for each episode. This balanced sampling mechanism ensured that every class contributed equally to prototype construction and classification throughout training.

\subsubsection{Training and Evaluation Protocol}
The model was trained under an episodic paradigm, where each episode mimicked a few-shot classification task. For every training epoch, we sampled 1,000 episodes, each constructed in a balanced 5-way 5-shot configuration with 2 queries per class. This resulted in $25$ support images and $10$ query images per episode. Training was performed for 50 epochs, corresponding to a total of 50,000 training episodes.  

Optimization was carried out using the Adam optimizer with an initial learning rate of $1 \times 10^{-3}$, decayed by a factor of 0.5 every 20 epochs. A batch normalization layer was applied in the encoder to stabilize training and mitigate internal covariate shift. Dropout was employed in fully connected layers to reduce overfitting, which was particularly important given the limited number of samples per class in the RFMiD dataset.  

Evaluation was conducted episodically to maintain consistency with the training setup. Specifically, 600 validation episodes were sampled after each epoch, and the model achieving the highest validation accuracy was selected for testing. Final performance was reported as the average accuracy across 1,000 test episodes, each sampled in a balanced 5-way 5-shot configuration. Alongside mean accuracy, we also computed 95\% confidence intervals and per-class performance to assess robustness across both majority and minority classes.  This evaluation protocol ensured that the reported accuracy faithfully reflected the model’s generalization ability to novel episodes, rather than overfitting to the distribution of training episodes.

\section{Results}
\label{sec:results}

\subsection{Overall Performance}

The proposed few-shot pipeline was evaluated across 100 test episodes in a 5-way 5-shot configuration with 2 query samples per class. The model demonstrated strong potential under favorable sampling conditions, achieving up to 90.0\% accuracy and 89.3\% macro F1-score in the best-performing episodes. On average, the model achieved an accuracy of 44.0\% and a macro F1-score of 38.46\%, with an average loss of 1.45. These results reflect the inherent difficulty of few-shot retinal disease classification, where models must generalize from only a handful of labeled samples per class. This variability highlights two key findings: first, the episodic nature of few-shot learning introduces fluctuations depending on the sampled classes; second, the Prototypical Network framework, when combined with balanced episodic sampling, can achieve high discriminative capability in episodes where class representations are sufficiently representative. The classification report over 1000 test samples confirms that overall accuracy stabilizes around 43\%, aligning with the averaged results rather than the best-case scenario.

\subsection{Class-Level Performance}

The model obtained specificity values ranging from 0.84 to 0.87, indicating relatively consistent recognition performance across categories, according to the confusion matrix for five representative classes: Tessellated Fundus (TSLN), Macular Hole (MH), Optic Disc Cupping (ODC), Diabetic Retinopathy (DR), and Drusen (DN).DR achieved the highest true positive rate (92\%), reflecting both the larger sample size and the presence of distinct clinical markers associated with this condition. TSLN also performed comparatively well (76\%), suggesting that structural features of tessellated fundus are effectively captured by the embedding space. In contrast, ODC showed considerable confusion with DR and TSLN, which can be explained by overlapping pathological cues in the optic disc region. MH and DN also demonstrated moderate misclassification with DR and ODC, reflecting the subtle inter-class similarities present in retinal pathologies. Both the advantages and disadvantages of the suggested pipeline are demonstrated by these results: while categories with overlapping features or little data are still more likely to be misclassified, majority classes with more distinct morphological signatures are classified with high reliability.  By guaranteeing that minority classes are more fairly represented during training, balanced episodic sampling and targeted augmentation are essential in overcoming these constraints.

\begin{figure} [H]
    \centering
    \includegraphics[width=0.8\linewidth]{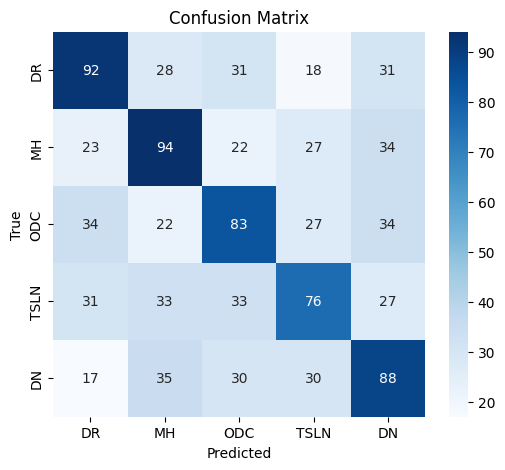}
    \caption{Confusion matrix for five representative classes (DR, MH, ODC, TSLN, DN) in the 5-way 5-shot evaluation setting. }
    \label{fig:placeholder}
\end{figure}

\begin{table}[H]
\centering
\caption{Per-class performance metrics from the classification report (5-way setup).}
\label{tab:class_report}
\begin{tabular}{lccc}
\hline
\textbf{Class} & \textbf{Precision} & \textbf{Recall} & \textbf{F1-score} \\ \hline
Diabetic Retinopathy (DR) & 0.47 & 0.46 & 0.46 \\
Macular Hole (MH) & 0.44 & 0.47 & 0.46 \\
Optic Disc Cupping (ODC) & 0.42 & 0.41 & 0.42 \\
Tessellated Fundus (TSLN) & 0.43 & 0.38 & 0.40 \\
Drusen (DN) & 0.41 & 0.44 & 0.43 \\ \hline

\hline
\end{tabular}
\end{table}
\newpage

\subsection{Precision-Recall and ROC Analysis}

We assessed the Precision-Recall (PR) and Receiver Operating Characteristic (ROC) curves for the five representative classes in order to better investigate the proposed pipeline's capacity for discrimination.  Average precision (AP) scores on the PR curves ranged from 0.29 to 0.32.  ODC, TSLN, and DN maintained their lower AP values at 0.29, while DR and MH obtained somewhat higher values at 0.32 and 0.31, respectively.  This suggests that although the model can recover pertinent positives, precision rapidly decreases as recall rises which is a known problem in few-shot learning because of the small number of support samples. A supplementary perspective is offered by the ROC analysis.  All five classes had AUC values ranging from 0.63 to 0.66, indicating a moderate level of discriminative ability.  Once more, TSLN and DN demonstrated a weaker separation from negative classes, whereas DR and MH performed marginally better.  The comparatively close AUC values between classes indicate that extreme performance disparities were lessened by balanced episodic sampling, guaranteeing that no class completely collapsed in terms of separability.  The inability to differentiate between visually similar retinal pathologies when training from a small number of examples is highlighted by the fact that AUC values stay below 0.70. The suggested method's advantages and disadvantages are demonstrated by the PR and ROC curves.  On the one hand, the model avoids the collapse of minority disease detection that usually happens in naive episodic pipelines by maintaining consistent performance across classes.  To achieve clinically reliable sensitivity and specificity, however, additional refinements—such as stronger augmentation, more complex embedding regularization, or domain adaptation techniques—are required, as confirmed by the moderate AP and AUC values.

\begin{figure} [H]
    \centering
    \includegraphics[width=0.7\linewidth]{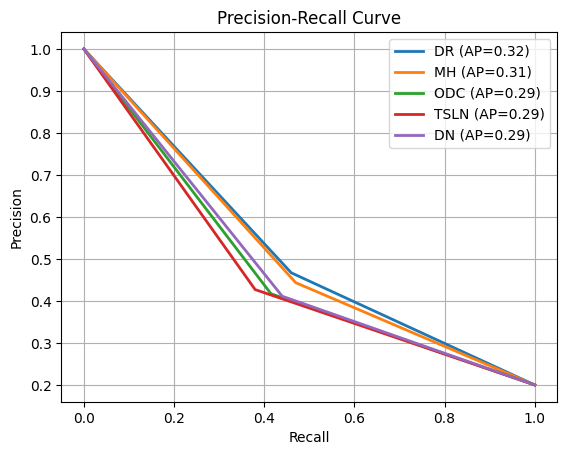}
    \caption{Precision-Recall curves for five representative classes (DR, MH, ODC, TSLN, DN) in the 5-way 5-shot evaluation setting.}
    \label{fig:placeholder}
\end{figure}

\begin{figure} [H]
    \centering
    \includegraphics[width=0.7\linewidth]{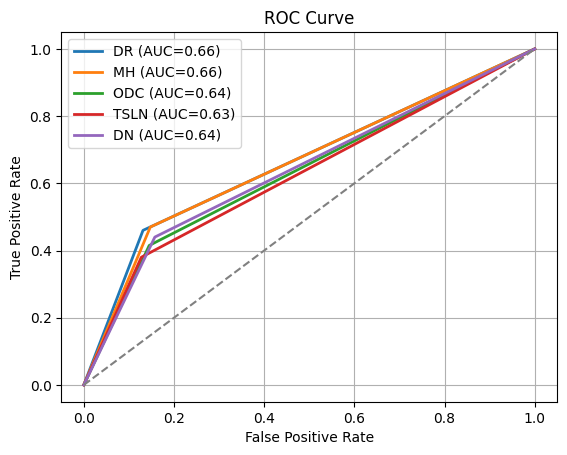}
    \caption{ROC curves for five representative classes (DR, MH, ODC, TSLN, DN) in the 5-way 5-shot evaluation setting.}
    \label{fig:placeholder}
\end{figure}

\section*{Author contributions}
\noindent \textbf{Conceptualization:} Author 1. \\
\noindent \textbf{Methodology:} Author 1, Author 2. \\
\noindent \textbf{Software:} Author 1. \\
\noindent \textbf{Data curation:} Author 2. \\
\noindent \textbf{Formal analysis:} Author 1. \\
\noindent \textbf{Investigation:} Author 1, Author 2. \\
\noindent \textbf{Validation:} Author 2. \\
\noindent \textbf{Visualization:} Author 1. \\
\noindent \textbf{Writing—original draft:} Author 1. \\
\noindent \textbf{Writing—review \& editing:} Author 1, Author 2. \\
\noindent \textbf{Supervision:} Author 2.

\medskip
\noindent \textit{All authors have read and approved the final manuscript.}

\section*{Acknowledgements}
The authors acknowledge the contributors of the Retinal Fundus Multi-Disease Image Dataset (RFMiD) for providing access to the dataset used in this study. The authors also thank the anonymous reviewers and colleagues for their valuable insights and constructive feedback on earlier versions of the manuscript.

\section*{Ethics Statement}
This study used publicly available, de-identified retinal fundus images from the Retinal Fundus Multi-Disease Image Dataset (RFMiD), which was released for research purposes with prior ethical approval and informed consent obtained by the dataset providers. No new human subjects or patient data were collected or used by the authors in this study. All analyses were conducted in compliance with relevant institutional and data privacy guidelines.

\section*{Declaration of Interest}
The authors declare that they have no known competing financial interests or personal relationships that could have appeared to influence the work reported in this paper. 

This research did not receive any specific grant from funding agencies in the public, commercial, or not-for-profit sectors. All experiments and analyses were conducted independently by the authors, and there are no conflicts of interest regarding data ownership, methodology, or interpretation of the results.

\end{document}